\documentclass[runningheads]{llncs}
\usepackage[T1]{fontenc}
\usepackage[utf8]{inputenc}
\usepackage{graphicx}
\usepackage{amsmath,amssymb} 
\usepackage{color}
\usepackage[width=122mm,left=12mm,paperwidth=146mm,height=193mm,top=12mm,paperheight=217mm]{geometry}

\usepackage{import}

\usepackage{tabularx}
\usepackage{booktabs}
\usepackage{multirow,array}
\usepackage{xspace}

\usepackage{hyperref}

\def\ie{\emph{i.e.}\xspace}
\def\eg{\emph{e.g.}\xspace}

\def\etal{\emph{et al.}\xspace}

\def\3drms{3DRMS\xspace}
\def\elas{ELAS\xspace}

\begin{document}
\pagestyle{headings}
\mainmatter

\title{Technical Report: Co-learning of geometry and semantics for online 3D mapping} 

\titlerunning{Technical Report: ECCV 2018 - 3DRMS Workshop Challenge}

\authorrunning{M. Carvalho, M. Ferrera, A. Boulch, J. Moras, B. Le Saux, P. Trouvé-Peloux}

\author{Marcela Carvalho$^{\dag}$, Maxime Ferrera$^{\dag}$, Alexandre Boulch, Julien Moras, Bertrand Le Saux, Pauline Trouvé-Peloux}

\institute{DTIS, ONERA, Universit\'e Paris Saclay F-91123 Palaiseau - France \\ $^{\dag}$ Equal Contribution}


\maketitle

\begin{abstract}
\textit{
This paper is a technical report about our submission for the ECCV 2018 3DRMS Workshop Challenge on Semantic 3D Reconstruction \cite{Tylecek2018rms}.  In this paper, we address 3D semantic reconstruction for autonomous navigation using co-learning of depth map and semantic segmentation. The core of our pipeline is a deep multi-task neural network which tightly refines depth and also produces accurate semantic segmentation maps. Its inputs are an image and a raw depth map produced from a pair of images by standard stereo vision. The resulting semantic 3D point clouds are then merged in order to create a consistent 3D mesh, in turn used to produce dense semantic 3D reconstruction maps. The performances of each step of the proposed method are evaluated on the dataset and multiple tasks of the \3drms Challenge, and repeatedly surpass state-of-the-art approaches. 
}
\end{abstract}

\begin{figure}[!ht]
\centering
\vspace{-5mm}
\includegraphics[width=0.95\linewidth]{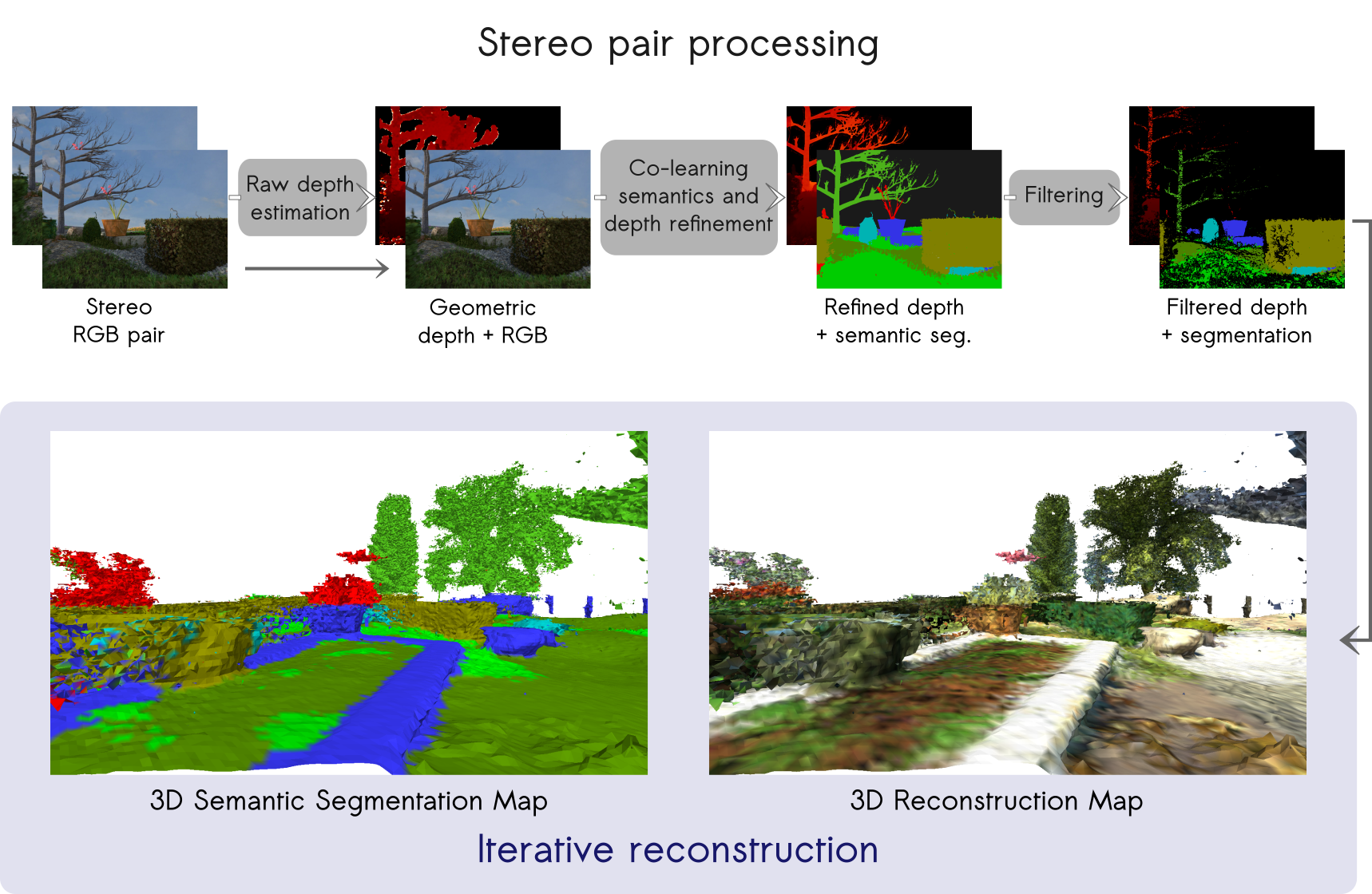}
\caption{Pipeline for generating geometric and semantic 3D reconstruction maps. 
}
\label{fig:pipeline}
\end{figure}

\section{Introduction}

Autonomous navigation is conditioned by the ability of sensing and analyzing the environment to take new decisions.
In this context, accurate 3D reconstruction and semantic understanding of the scenes are critical.
Indeed, building a 3-Dimensional (3D) map of the scene including semantic information allows to plan future trajectories accordingly to the tasks to perform.

Over the past years, improvements on data acquisition techniques and processing made possible reconstructing 3D scenes in multiple ways.
On the one hand, active sensors are now mature technology and some variants gain special attention, like LIDARs, which produce dense and reliable point clouds~\cite{cole-icra06-lidar-slam}; and RGB-D sensors that generates corresponding depth maps which can be combined to scene reconstruction~\cite{ElasticFusion}.
On the other hand, passive approaches like Structure-from-Motion (SfM) are also commonly adopted to recover 3D relations between points and objects from a set of 2D images. In SfM, we distinguish offline methods from online methods. Offline variants, also denoted as photogrammetry, usually exhaustively process all data before global reconstruction. In opposition, online approaches handle information incrementally to perform reconstruction while data are being acquired, as in Simultaneous Localization and Mapping (SLAM) and in Visual Odometry (VO). However, these latter techniques usually rely on geometric features and not on semantic information, though it is an important feature to perform more specialized and complex navigation tasks.

In this work, we present a new approach to jointly learn geometry and semantics for 3D mapping. The proposed pipeline consists of two steps, corresponding to different levels of data aggregation (Fig.~\ref{fig:pipeline}). First, at image level, a multi-task network estimates a depth map and a semantic segmentation map. Then, these geometric and semantic features are accumulated into a global representation 
where the semantic mesh of the scene is extracted from the 3D representation, which allows scene understanding and planning of further actions. 

In details, the main contributions of this paper are the following. The first key point is the joint use of geometric and machine learning approaches. As illustrated in Figure(~\ref{fig:pipeline}), a raw depth map is estimated from a pair of images using stereo and then is refined through a convolutional neural network. A second key point is the co-learning of depth and semantic segmentation from the raw depth map and an RGB images. Hence the proposed network performs multiple tasks at once, with mutual benefit. We show that this approach leads to better performances than independent predictions of depth and semantic segmentation. Finally, on the contrary to global, offline reconstruction methods, our approach is incremental and hence is conceptually compatible with autonomous navigation and robotics.

The paper is organized as follows: section~\ref{sec:related} presents works related to the problem, section~\ref{sec:method} describes our semantic reconstruction pipeline and finally section~\ref{sec:results} evaluates our method with quantitative and qualitative results on the \textit{3D Reconstruction Meets Semantics 2018} (\3drms) Challenge dataset, which contains series of stereo sequences generated over a simulated garden.

\section{Related work}
\label{sec:related}

\textbf{Perception for autonomous navigation} has been a great topic of interest in the last two decades.  As cameras became cheap and easy to embed while still offering rich information, vision-based SLAM methods grew more and more popular~\cite{ORB-SLAM,SVO-2}. SLAM allows a robot to localize itself with respect to the environment. Either this environment is unknown, and its 3D structure is simultaneously estimated, or the environment is already known, and a previously built map can be used~\cite{lynen2015getoutofmylab,schneider2018maplab}. In the latter case, such maps can be obtained by regular SLAM methods, \ie. building the map of the environment and then using it for self localization. Maps can also be built offline by SfM algorithms such as Colmap~\cite{Colmap} or OpenMVG~\cite{openMVG} before being used for real-time localization. All these approaches for offline or online map construction, take only the geometric structure of the scene into account. However, a few works proposed to also benefit from semantic information, yielding in \textit{semantic SLAM}~\cite{civera-iros11-semantic-slam}. Indeed, this allows to get better maps and increase the localization reliability~\cite{semfusion,SchoenbergerCVPR18semvloc}. Using RGB-D data, a pipeline using random forests for creating semantic maps in 2D and 3D was proposed in~\cite{hermans-icra14-3D-semantic-mapping-and-reconstruction}. More recently, \cite{lingni17iros} applied joint learning with neural networks over multiple RGB-D views to generate better 2D semantic maps, but did not reconstruct corresponding 3D models. With respect to all these approaches, ours offers a functional pipeline from 2D images to 3D reconstruction with semantics. With respect to the latter ones, semantics and geometry have a better integration directly in the network.

The \textbf{joint use of geometry and semantics} has been investigated in the previous edition of the \3drms challenge~\cite{sattler-isr18-3DRMS}. The dual objective was 3D geometry reconstruction and semantic classification. The proposed baseline links Colmap~\cite{Colmap} and SegNet~\cite{badrinarayanan2015segnet} (for 2D classification, then projected in 3D). Both entries~\cite{guerry-17iccvw-snapnet-R,taguchi-chen-3drms} used semantics during the reconstruction to filter out outliers, but with worst performances than the baseline. With respect to these approaches, we propose to learn the fusion of semantics and geometry through a multi-task network.

Fully Convolutional Networks (FCNs)~\cite{long2015fcn,badrinarayanan2015segnet,Ronneberger2015} have been widely used for many tasks in computer vision. In brief, they are dense prediction methods which intend to assign information back onto the original pixels positions.
\textbf{Semantic segmentation} is a common domain of application for such dense prediction networks.
 We focus here on the approaches which benefit from geometric information. FuseNet~\cite{hazirbas2016fusenet} uses two interlaced encoders and a single decoder for semantic segmentation from RGB-D data.
Alternatively, in~\cite{audebert-16accv-residual-correction}, the authors introduce \textit{residual fusion} using a small network to merge the outputs of two SegNets applied to different sensor modalities. A finer (though more complex) approach, 3D graph neural network~\cite{qi20173d}, consists in considering information extracted from the local 3D graph of adjacency and using it in the segmentation network. \cite{guerry-17iccvw-snapnet-R} proposed 3D-consistent data augmentation to incorporate the geometry directly in the training set. Among all these approaches, the one which has most in common with ours is FuseNet~\cite{hazirbas2016fusenet}, since they share solving the fusion problem by a highly-integrated network. However, our network goes beyond simple fusion, and address a multi-task problem, with semantic segmentation and depth adjustment.


FCNs have also been applied to other tasks such as monocular \textbf{depth prediction}~\cite{Eigen2015,xu2017multi,kendall2017uncertainties,Carvalho2018icip}. These techniques exploit spatial correlation based on structured information (\eg, linear perspective, textures) 
to produce reliable depth maps from 2D scenes. To cite but a few, we then focus on approaches with open-source code. Based on SegNet, Laina~\etal~\cite{laina2016deeper} exploit residual connections~\cite{he2016deep} and fast up-projection blocks. In D3-Net~\cite{Carvalho2018icip}, the network consists on a densely connected encoder~\cite{huang2017densely} and a U-Net like decoder structure to predict refined depth estimation. With respect to these approaches, our method also uses depth from geometry as an input, and refines the 2D depth map using semantic constraints, which yields in better depths than with stereo or monocular prediction.


\section{Proposed approach}
\label{sec:method}
As 
presented in figure~\ref{fig:pipeline}, our method is composed of two computation levels: 
depth and semantic maps generation; 3D data accumulation for surface reconstruction. These tasks are combined sequentially and result in an accurate method for 3D scene reconstruction. From beginning to end, we use a stereo sequence to produce a semantic mesh. 

Our main idea is to learn jointly the depth and the semantic segmentation in a multi-task deep neural network framework. Besides, we also benefit  from geometric depth estimation methods. Indeed, raw depth map estimated from a pair of stereo images with geometric approach are used as inputs of the multi-task network. In the following, we describe in details the four sub-tasks of Fig.~\ref{fig:pipeline}.


\subsection{Depth estimation  \label{sec:geomdepth}}
The first step of the proposed 3D reconstruction pipeline consists in estimating depth maps from stereo views. In brief, the calibration of stereo cameras allows estimating the relative pose of the right camera with respect to the left one, as well as their distortion parameters.  Using these informations, the left and right images may be undistorted and rectified in order to be aligned.  Once aligned, the depth of corresponding points in both images can be estimated from the known baseline between the cameras, their focal length and the disparity between the two points.

Two different stereo matching algorithms have been tested to compute disparity maps: \elas \cite{Geiger10elas} and SGBM \cite{Hirschmueller08sgbm}.  \elas is a probabilistic method based on the triangulation of robust matches to create reliable support points.  These support points then serve as priors for the disparity search and allows the computation of a Maximum-A-Posteriori (MAP) estimate over the remaining pixels. \elas was tested using the LIBELAS implementation, without post-processing. Indeed, some post-processing is commonly applied in order to get better and more dense disparity maps. However, the neural networks might be more prone to handle raw disparity maps than human-interpretable ones. Otherwise, SGBM is a semi-global method which estimates disparity by minimizing an energy function made of the Sum of Absolute Distances (SAD) over a local window and a smoothness term. SGBM was tested using its OpenCV implementation and no post-processing were applied.


Their respective accuracy results on the training sequences 0001 and 0224 of the \3drms dataset are displayed in table \ref{table:error_measurements}. SGBM gives slightly better results than \elas, so the depth maps produced by SGBM are the ones we choose to use in the proposed pipeline.  

\begin{figure}[t]
\includegraphics[width=\linewidth]{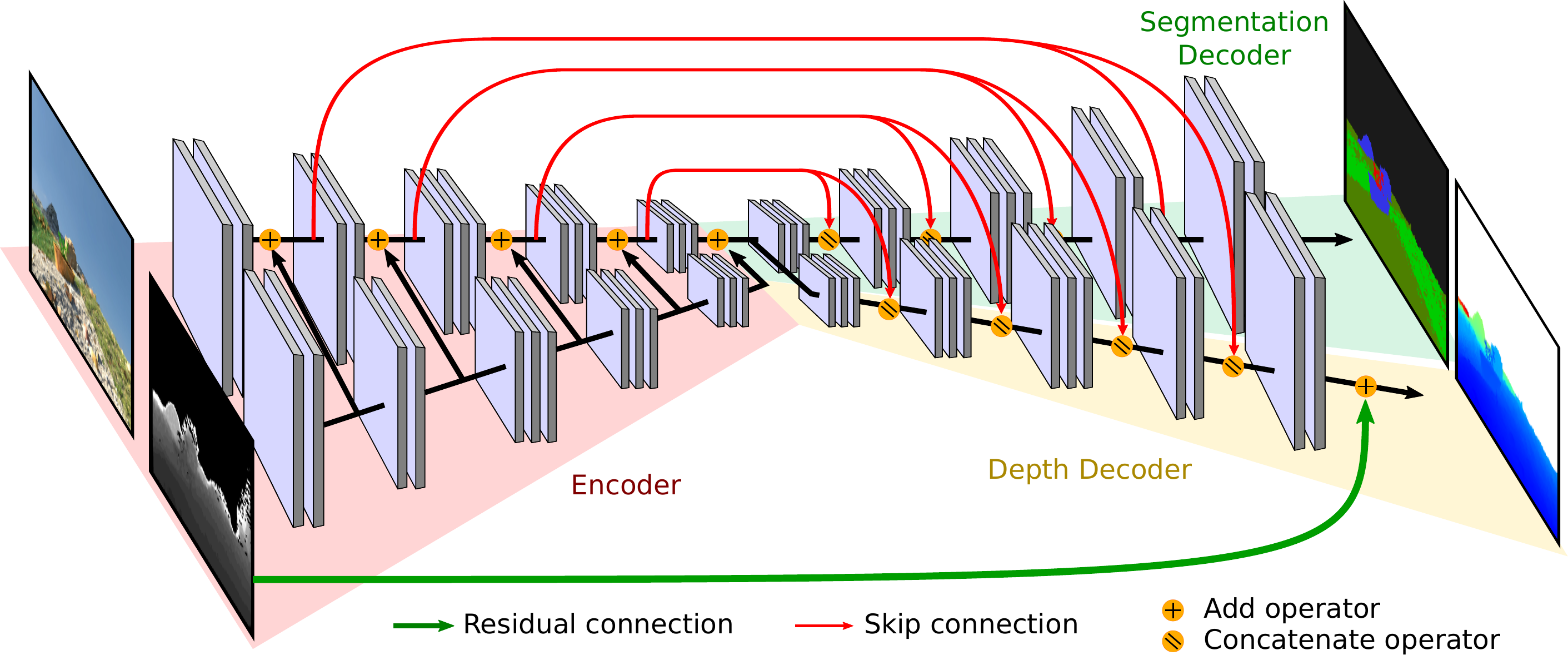}
\caption{Multi-task Network architecture.}
\label{fig:network}
\end{figure}

\subsection{Semantic segmentation and depth enhancement \label{sec:segsem}}

The task at hand here is the reconstruction of a semantic mesh of the given scene.
Hence, the objective is twofold: reconstruct the geometry of the scene (3D localization of the mesh vertices) and identify semantics (attach a label to each mesh element). 

Though, the geometric depth estimation previously described is not sufficient for global surface reconstruction. Indeed, the small baseline of the stereo sensor might lead to huge estimation errors. Hence, a refinement step is needed to produce better depth maps. As the geometric errors mostly occur on edges, using the RGB image as an additional information would lead the network to produce sharper edges. Besides, as shown in \cite{hazirbas2016fusenet,qi20173d}, semantic segmentation benefits from both RGB images and depth maps.

These considerations motivate the proposed approach of a multi-task fully convolutional neural network for a joint prediction of depth and semantic segmentation. The proposed architecture was inspired in FuseNet~\cite{hazirbas2016fusenet} and is presented on figure~\ref{fig:network}. The Multi-task Network has an encoder-decoder structure, with two branches for the encoder, and 2 independent branches for the decoder (one for semantics, one for depth estimation). Contrarily to the original implementation of FuseNet, we add skip connections between the encoder and decoder parts to improve spatial information flow over the network. Branches in the contractive part take the RGB and raw depth inputs respectively and as feature maps are generated, they are melt from the depth branch to the RGB input branch. Also, depth refinement is performed in a residual manner, adding the correction to the input raw map.

\subsection{Filtering}


Even though depth is enhanced using the multi-task network, a few errors remain when an object occludes another. In this case, the network tends to smooth the transition between objects and overlook small details (such as tree leafs for example). To avoid unwanted outliers in later stages of the 3D reconstruction, we apply the following filtering operations. First, points labeled as sky are removed. Second, points from uncertain object borders are identified and removed. These borders correspond to transitions between objects at different depths, so we compute the gradient of depth over the image and remove all pixels for which the gradient norm is greater than a fixed threshold (empirically set to $0.05$). Finally, we also test in the experiments an additional filter: erosion, or removal of the neighbors of a point considered as an outlier. As we will show, this produces more precise but less complete reconstructions.


\subsection{Iterative 3D map construction}

The 3D reconstruction module is based on Truncated Signed Distance Function (TSDF) modeling. This technique estimates a scalar field which represents the approximate distance of every points in the 3D space to the nearest surface. In practice, the field is estimated over a 3D discretization of the world and only close to the surfaces. The distance estimates are signed: positive outside of the object and negative for the inside. Hence, the zero crossing is an implicit representation of the surfaces of the objects present in the scene and a \textit{Marching Cube} algorithm is used to recovers the mesh.
The TSDF implementation used in this paper is based on OpenChisel~\cite{Klingensmith-2015-5990}. In order to estimate the distance field, the 3D space is discretized into voxels and the filtered depth maps are integrated into the TSDF according to the poses of the camera.  The depth maps are first clipped in order to only process 3D points within a clipping range distance from the camera (in practice from 0.5 to 5-10 meters).  In addition to distance estimations, we also add semantic classification fusion.  Thus the module can take as a new input, either the label image resulting from classification or directly the classification scores (cf. Section~\ref{sec:segsem}).  These semantic inputs are processed in the same way as the depth maps, that is the voxels integrate the semantic scores in addition to the distance-to-nearest-surface values.  When all the frames have been integrated, a filter removes the voxels which do not contain accurate enough distance values.  For each remaining voxel, the semantic label is selected as the one with highest score.  In practice, the voxel grid resolution is set to 3cm.  The mesh is finally generated by applying \textit{Marching Cube} over this voxel grid.


\section{Experiments results}
\label{sec:results}

In this section, each step of the semantic reconstruction pipeline is evaluated on the \3drms dataset.
The data consists in four training sequences with ground truth and a test sequence for challenge evaluation (for which the ground truth remains undisclosed to participants).
We further divide the training set in train and validation to present evaluation scores and comparable visual results. Precisely, we created two folds from the training data: fold 1 with training scenes from sequences 0128, 0160, 0224 and testing scenes from sequence 0001; and fold 2 with training scenes from sequences 0001, 0128, 0160 and testing scenes from sequence 0224.

In the following, semantic segmentation (Section~\ref{sec:res-semseg}), depth estimation (Section~\ref{sec:res-depth}) and global 3D reconstruction (Section~\ref{sec:res-reconst}) are  evaluated on this dataset and compared to state of the art approaches. Metrics used for comparison are presented in tables \ref{table:error_measurements}.

\begin{table}[t]
    \caption{\label{table:error_measurements}Error measurements adopted to evaluate semantic segmentation (left) and  depth estimation performances (right). $P$ corresponds to the predictions and $GT$ the ground truth. $C$ is the number of classes. Variables $d_i$ and $\hat{d_i}$ are the ground truth and prediction respectively, and $N$ is the total number of pixels.}
\centering
\footnotesize
\begin{minipage}{0.47\linewidth}
\centering
\begin{tabular}{lc}
\toprule
Metric  & Definition\\
\midrule
Overall Acc. (OA) & $\frac{P \cap GT}{|GT|}$ \\
Average Acc. (A.Acc.) & $\frac{1}{C} \sum_{i=1}^C \frac{P_i \cap GT_i}{|GT_i]}$ \\
Average IoU (A. IoU) & $\frac{1}{C} \sum_{i=1}^C \frac{P_i \cap GT_i}{P_i \cup GT_i}$ \\
\bottomrule
\end{tabular}

\end{minipage}
\hfill
\begin{minipage}{0.52\linewidth}
\centering
    \begin{tabular}{@{}lc@{}}\toprule
    Metric & Definition \\
    \midrule
    Abs. error & $\frac{1}{N}\sum_{i=0}^{N}\frac{|d_{i}-\hat{d}_{i}|}{d_{i}}$\\
    RMS & $\sqrt{\frac{1}{N}\sum_{i=0}^{N}(d_{i}-\hat{d}_{i})^2}$\\
    \bottomrule
    \end{tabular}

\end{minipage}
\end{table}

\begin{table}[t]
\caption{Comparison of semantic segmentation results on the \3drms 2018 dataset using state-of-the-art segmentation networks and the proposed Multi-task Network.}
\label{tab:semseg}
\tiny
\centering
\begin{tabular}
{@{}l@{\hspace{2mm}}l@{\hspace{2mm}}l@{\hspace{2mm}}c@{\hspace{2mm}}c@{\hspace{2mm}}c@{\hspace{2mm}}c@{\hspace{2mm}}c@{\hspace{2mm}}c@{\hspace{2mm}}c@{\hspace{2mm}}c@{\hspace{2mm}}}
\toprule
\multicolumn{3}{c}{Methods} & \multicolumn{3}{c}{Test on 0001} && \multicolumn{3}{c}{Test on 0224} \\
\cmidrule{4-6}\cmidrule{8-10}
      & Input & Output & OA & A. Prec. & A. IoU && OA & A. Prec & A. IoU \\
\midrule
\textit{Baselines}\\
U-Net            &RGB  & S& 0.9068 & 0.8286 & 0.7012 && 0.9054 & 0.7496 & 0.6395 \\
FuseNet          &RGB  & S& 0.9091 & 0.8577 & 0.7371 && 0.9311 & 0.8038 & 0.7169 \\
\midrule
\textit{Multitask refinement}\\
Multi-task Net.   &RGB+D$_{\elas}$& D+S &0.9363 & 0.8916 & 0.7943 && 0.9277 & 0.7952 & 0.7107 \\
Multi-task Net.   &RGB+D$_{SGBM}$& D+S &0.9411 & 0.8965 & 0.7980 && 0.9303 & 0.8017 & 0.7195 \\
\bottomrule
\end{tabular}
\end{table}

\subsection{Semantic segmentation \label{sec:res-semseg}}

\begin{figure}[t]
  \centering
  \def\svgwidth{1.2\textwidth}
  \scalebox{0.7}{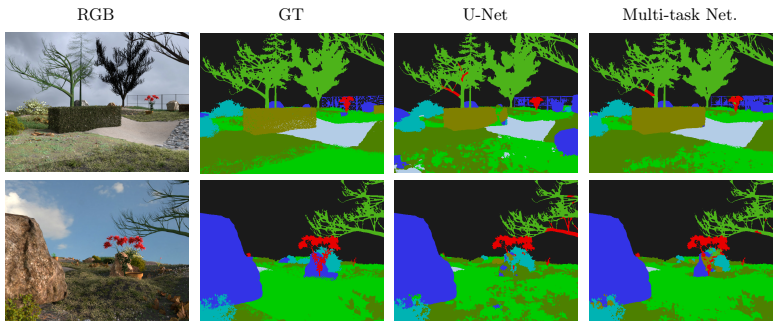}
  \vspace{5mm}
  \caption{Comparison of the semantic segmentation maps generated by U-Net trained on RGB images; and the proposed multi-task network architecture trained on RGB and raw depth images on input and extra refined depth on output.
  }
  \label{fig:semantic_map}
\end{figure}

Semantic segmentation from 2D images is one of the tasks of the \3drms Challenge.
Our architecture is evaluated against the two state-of-the-art approaches: U-Net~\cite{Ronneberger2015} and FuseNet~\cite{hazirbas2016fusenet}.  Performances are computed according to the metrics presented in Table~\ref{table:error_measurements}-left. Results are presented in Table~\ref{tab:semseg}. It shows a clear improvement of the performance of semantic segmentation when using the proposed multi-task network. The best configuration is multi-task training using the raw depth map from SGBM. In Fig.~\ref{fig:semantic_map}, examples of semantic segmentations of some 2D images from the dataset are displayed. It shows that co-learning enforces consistency with respect to the 3D structure. Indeed, neighbor pixels with the same depth (\ie also close to each other in 3D) tend to get the same semantic label.

\subsection{Depth estimation  \label{sec:res-depth}}

The quality of 3D reconstruction highly depends on the estimation of an accurate depth map. In this paper, we propose to generate a precise depth map by refining a raw one obtained with a stereo pair. This process is one of the tasks of our multi-task network. In the following, we compare the performances of depth estimation using traditional stereo methods such as \elas\cite{Geiger10elas} and SGBM \cite{Hirschmueller08sgbm} with the performances of the  refined depth estimates. We also evaluate the performance of a state-of-the-art single-image depth estimation approach, referred to as D3-Net~\cite{Carvalho2018icip}. \\

\begin{figure}[t]
\centering
\begin{tabular}{cc}
\includegraphics[width=0.44\linewidth]{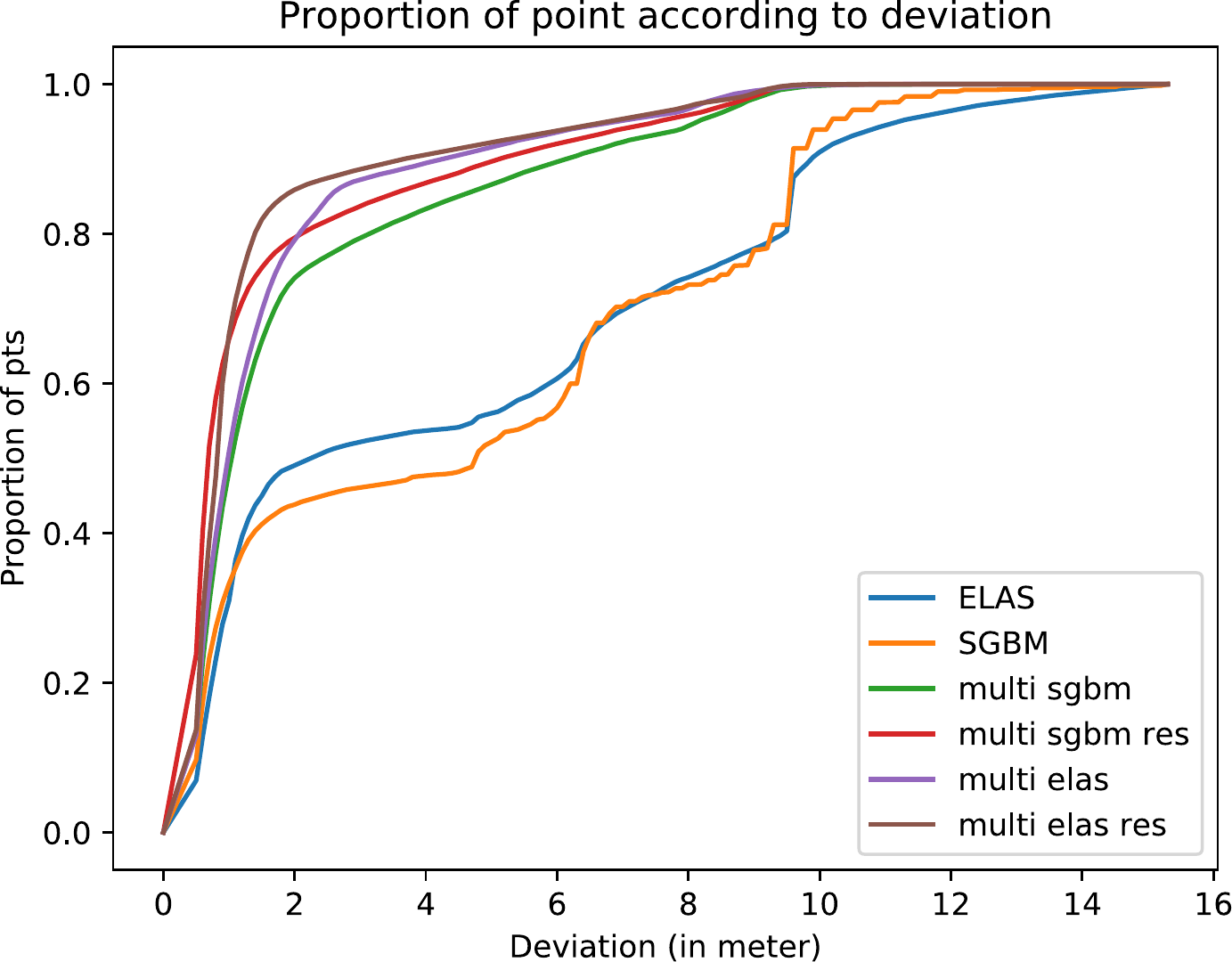}
&
\includegraphics[width=0.45\linewidth]{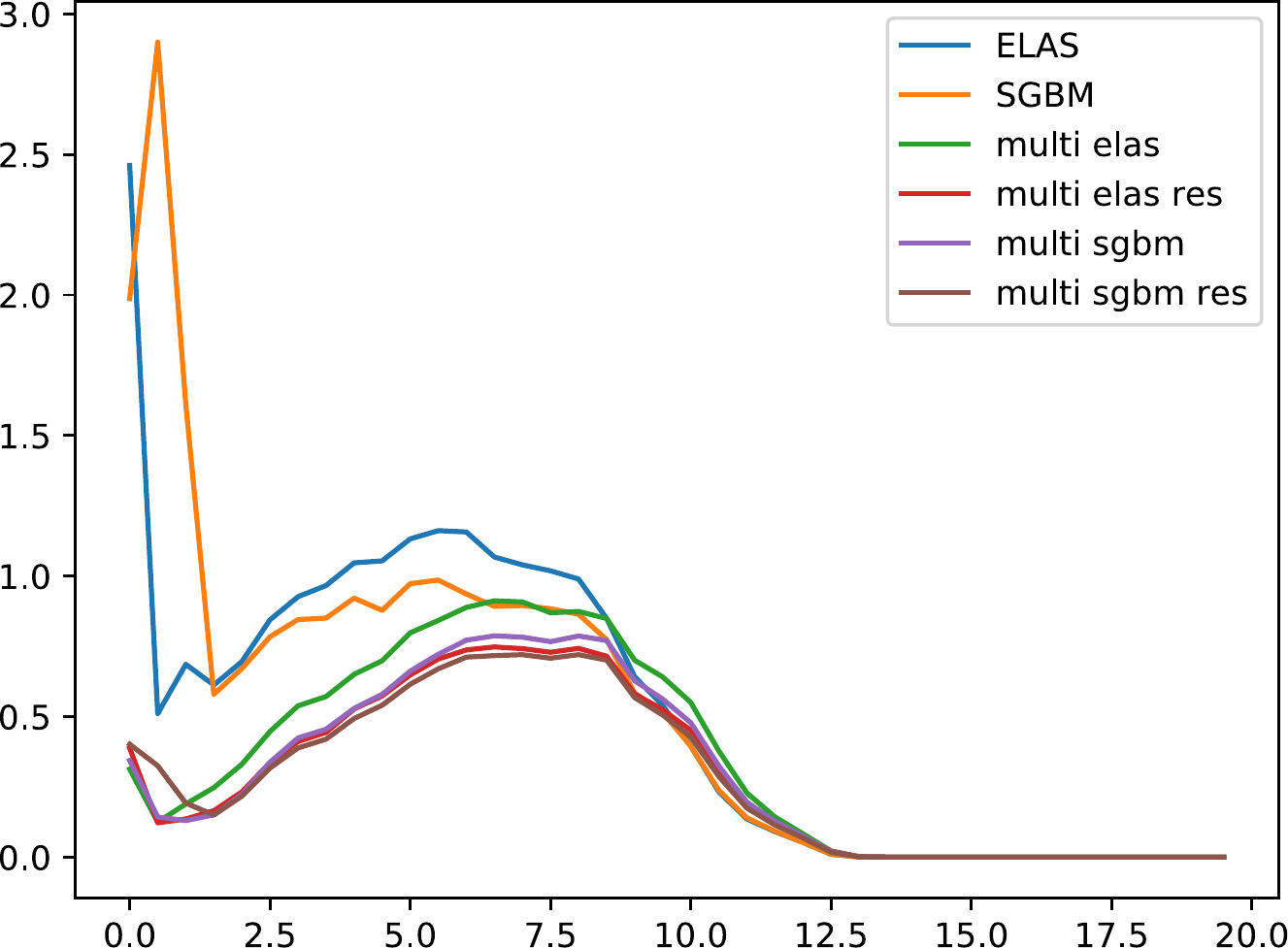}\\
(a)&(b)
\end{tabular}
\caption{Comparison of performances of geometric methods (\elas and SGBM) and the proposed approach, depth estimation through co-learning: (a) Proportion of 3D points with deviation less than a given value; (b) RMS of the depth estimates with respect to the ground truth distances.}
\label{fig:prop_and_RMS_depth}
\end{figure}

The various depth map predictions are first compared using standard error measurements previously used for the same purpose~\cite{li2015depth,Eigen2015} and defined in Table~\ref{table:error_measurements}(right).
We also provide the proportion of points with a deviation less than a given value in Fig.~\ref{fig:prop_and_RMS_depth}(a) and the RMS function with respect to the ground truth distance in Fig.~\ref{fig:prop_and_RMS_depth}(b).
Several conclusions can be drawn from this. First, refinement of the geometric depth map using a multi-task neural network highly improves the depth estimation accuracy. Indeed for geometric approaches, only 40~\% of the points have a deviation lower than 2m, while it reaches 80~\% using the proposed multi-task approaches.  One can note that improvement is specifically significant for small depth range, between 0 to 5m, which is crucial for safe autonomous navigation. Second, all multi-task learning approaches show similar results, with slightly better performances when the raw input comes from SGBM. Furthermore, our tests also show that using a state-of-the-art FCN for single-image depth estimation outperforms the purely geometric approaches according to these standard, global metrics. As discussed in the following, this result can be explained by a better depth map segmentation obtained by deep learning approach. 

\begin{table}[t]\centering 
    \caption{Comparisons of error metrics for depth estimation using geometric, D3-Net depth prediction and multi-task learning on the \3drms 2018 dataset. }
    \label{table:depth_estimation_results2}
    \scalebox{0.7}{
    \begin{tabular}    
{@{}l@{\hspace{2mm}}l@{\hspace{2mm}}l@{\hspace{3mm}}r@{\hspace{1mm}}r@{\hspace{3mm}}r@{\hspace{3mm}}r@{\hspace{1mm}}r@{\hspace{3mm}}}\toprule
\multicolumn{3}{c}{Methods} & \multicolumn{2}{c}{Error$\downarrow$ \textbf{Test 0001}}  & & \multicolumn{2}{c}{Error$\downarrow$  \textbf{Test 0224}} \\
    \cmidrule{4-5} \cmidrule{7-8} 
    &Input & Output& rel & rms &&  rel & rms   \\  \midrule
    
    \textit{Geometric} \\
    \elas                 &RGBx2 &D & 0.526 & 2.140 && 0.444 & 1.993  \\
    SGBM                 &RGBx2 &D & 0.518 & 1.801 && 0.439 & 1.745  \\
    \hline
    \textit{Mono image} \\
    D3-Net Mono			 &RGB   &D   & 0.145 & 0.755 && 0.110 & 0.477  \\
    \hline
    \textit{Refinement} \\
	FuseNet         &RGB+D$_{SGBM}$ &D   & 0.057 & 0.395 && 0.074 & 0.454  \\
    Multi-task Net.          &RGB+D$_{\elas}$ &D+S & 0.079 & 0.410 && 0.066 &  0.414 \\
	Multi-task Net.      &RGB+D$_{SGBM}$ &D+S & 0.082 & 0.394 && 0.089 & 0.436  \\

    \bottomrule
    \end{tabular}
    }
\end{table}

\begin{figure}[t]
\includegraphics[width=\linewidth]{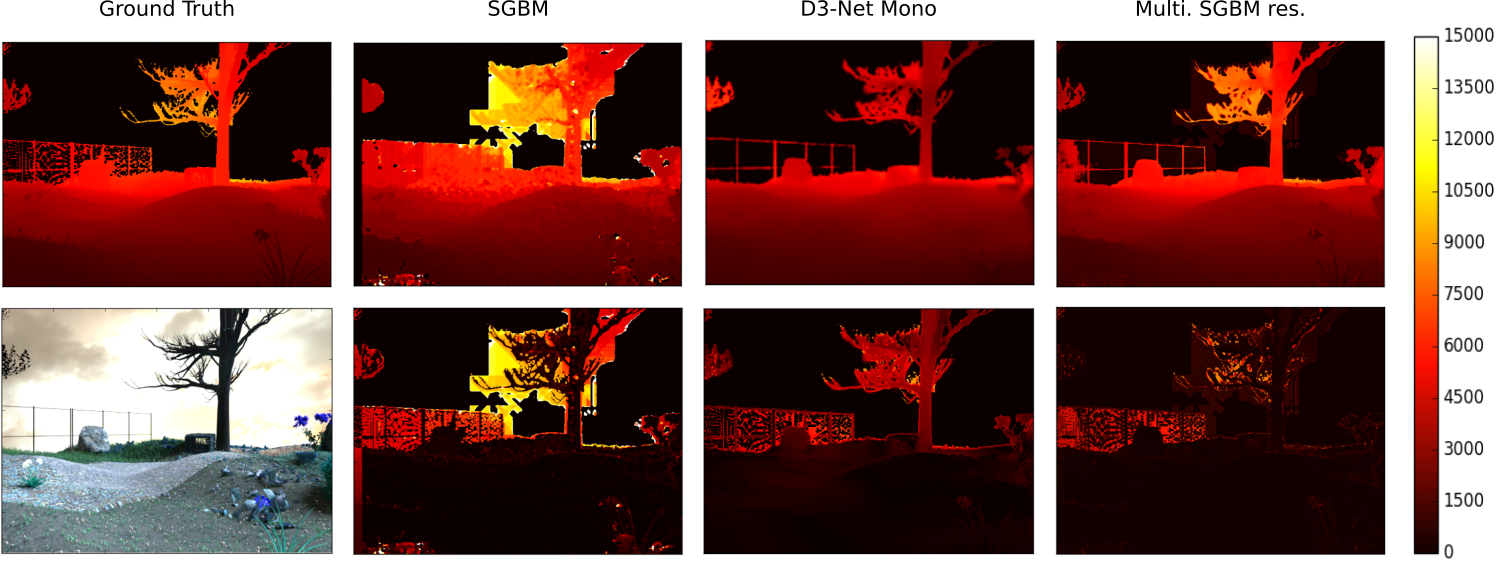}\\
\includegraphics[width=\linewidth]{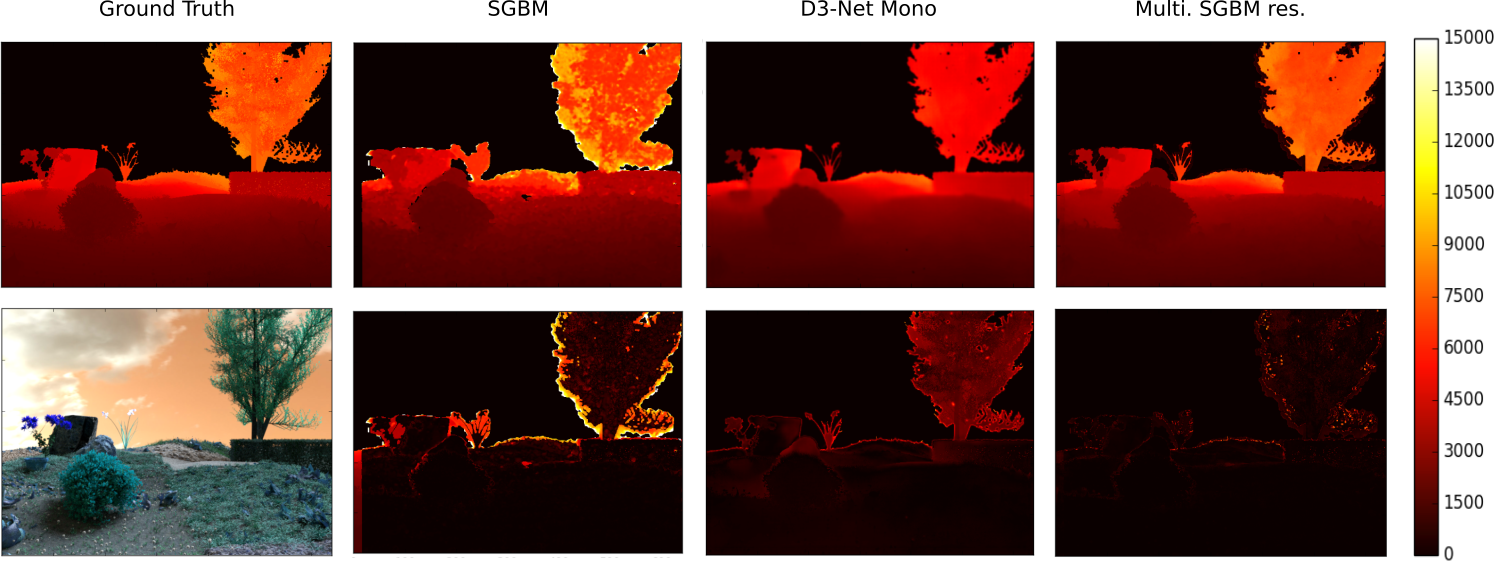}
\caption{Comparison of various depth estimation approaches: geometric approach (SGBM); depth prediction (D3-Net); and the proposed multi-task network (last column). \textit{First row / Last row: depth maps / error maps in mm.}}
\label{fig:deep_depth}
\end{figure}

Figure~\ref{fig:deep_depth} shows examples of depth maps obtained with the various geometric or multi-task approaches. 
A geometric method such as SGBM results in accurate depth estimates but with a low quality segmentation of the depth map. On the contrary, a deep learning approach such as D3-Net shows an excellent depth segmentation, but produces biased depth values. Finally, the proposed approach which benefits from both geometrical and deep learning techniques shows the best results both in terms of accuracy and quality of depth segmentation. 


\subsection{Reconstruction  \label{sec:res-reconst}}

The reconstruction is performed with OpenChisel~\cite{Klingensmith-2015-5990}.  
In this section, we present the final results of the reconstruction for both test sets 001 and 224 of the \3drms dataset.
We evaluate the geometric quality of the reconstruction according to the depth map filtering strategy.
We also provide quantitative and qualitative results on the semantics in 3D.

\begin{table}[t]
\caption{Error measurements used for geometric reconstruction quality estimation.}
\label{tab:metricsRec}
\centering
\begin{tabular}{l@{\hspace{4mm}}l}
\toprule
Metric & Definition\\
\midrule
Average distance for A $\rightarrow$ B       & $\frac{1}{|A|} \sum_{p \in A} \min_{q \in B}{d(p,q)}$ \\
Accuracy \% < 5cm for A $\rightarrow$ B      & $\frac{100}{|A|}\sum_{p \in A} 1_{\min_{q \in B}{d(p,q)} < 0.05}$ \\
Completeness dist 90\% for A $\rightarrow$ B & $\min_{d} \frac{100}{|A|}\sum_{p \in A} 1_{\min_{q \in B}{d(p,q)} < d} = 90$\\
\bottomrule
\end{tabular}

\textit{Note: $1$ is the indicator function.}

\end{table}

\begin{table}[t]
\caption{Evaluation of the reconstruction on the \3drms dataset.}
\label{tab:3Deval}
\scriptsize
\centering
\begin{tabular}{ll@{\hspace{2mm}}cc@{\hspace{2mm}}c@{\hspace{2mm}}cc@{\hspace{2mm}}c@{\hspace{2mm}}cc@{\hspace{2mm}}c@{\hspace{2mm}}cc}\toprule
&&\multicolumn{5}{c}{Full scene} && \multicolumn{5}{c}{Cropped scene z=1m}\\
\multicolumn{2}{c}{Filtering method} & \multicolumn{2}{c}{GT $\longrightarrow$ Recons.} && \multicolumn{2}{c}{Recons. $\longrightarrow$ GT} && \multicolumn{2}{c}{GT $\longrightarrow$ Recons.} && \multicolumn{2}{c}{Recons. $\longrightarrow$ GT}\\
\cmidrule{3-4} \cmidrule{6-7} \cmidrule{9-10} \cmidrule{12-13} 

 &range & Av.  & Complet. && Av.  & Acc. && Av. & Complet. && Av. & Acc.\\
       &      & dist. & dist. $90\%$  && dist. & \% < 5cm &&  dist. & dist. $90\%$  && dist. & \% < 5cm \\
\midrule
No filtering       &  5m & 0.061 & 0.145 && 0.201 & 32.2\% && - & - && - & - \\
                   & 10m & 0.061 & 0.164 && 0.311 & 20.9\%  && - & - && - & - \\
Gradient           &  5m & 0.077 & 0.208 && 0.037 & 77.6\% && 0.042 & 0.102 && 0.027 & 86.3\% \\
                   & 10m & 0.058 & 0.156 && 0.047 & 70.5\% && 0.043 & 0.109 && 0.031 & 83.9\% \\
Gradient           &  5m & 0.128 & 0.427 && 0.024 & 87.2\% && 0.052 & 0.134 && 0.020 & 90.7\% \\
 + Erosion                  & 10m & 0.113 & 0.356 && 0.028 & 85.1\% && 0.052 & 0.134 && 0.022 & 89.8\% \\
\bottomrule
\end{tabular}
\end{table}

\subsubsection{Geometric reconstruction}

As defined in~\cite{sattler-isr18-3DRMS}, the quality of the reconstruction can be evaluated from two points of view.
First, each point of the ground truth must be close to a point of the reconstructed scene, this is the completeness of the reconstruction, \ie. it express how well the whole scene has been discovered and reconstructed.
Second, each point of the reconstruction must be close to a point of the ground truth, this is accuracy.
The accuracy aims at evaluating how well the reconstruction fits to the ground truth.
In practice a good reconstruction is a compromise between completeness and accuracy; 
filling the space with points would improve the completeness while selecting only few points,  well positioned, would improve the accuracy.

\noindent We use the following metrics for quantitative results (definition in table~\ref{tab:metricsRec}):
\begin{itemize}
\item \textit{from ground truth to reconstruction}: the average distance of GT point to the mesh, and the completeness (the distance $d$ such that $90\%$ of the GT points are at distance less than $d$ to the reconstruction).
\item \textit{from reconstruction to GT}: the average distance of mesh vertices to GT, and the accuracy (percentage of vertices at distance less than 5cm to the GT)
\end{itemize}
We compute these metrics using CloudCompare~\footnote{CloudCompare: \url{https://www.danielgm.net/cc/}}.
For an easier readability, we restrict the numbers to the test set 001, results on the 224 scene being consistent the previous ones, the prediction method if the method using residual depth output with SGBM data as input.

We compare the effect of the filtering policies.
We made reconstruction experiments with the two filtering methods and the baseline (no filtering) for OpenChisel clipping range $r$ in $[5,10]$ meters.

Table~\ref{tab:3Deval} presents these results for ranges 5m and 10m.
The results are first computed with the full scene ground truth (including complete trees) and then with a cropped ground truth at 1m height (corresponding to the use case of autonomous lawnmower).
As expected, giving all the points OpenChisel leads to better completeness but produce a lot of reconstruction artifacts, in particular at transitions between objects or sky.
Filtering these points based on gradient produce much better results according to outlier production while ensuring a good completeness.
On the opposite, the harder filter given by the gradient and an erosion improves even more the accuracy but drastically reduced the number of 3D points (\eg top of the bushes and topiaries) which impacts the completeness.
Better performances are achieved using a a cropped ground truth.
This is mostly due to the small baseline of the stereo images and the ground view,
leading to missing or uncertain tree reconstruction.

\begin{figure}[t]
\centering
\begin{minipage}{0.48\linewidth}
\centering
\includegraphics[width=\linewidth]{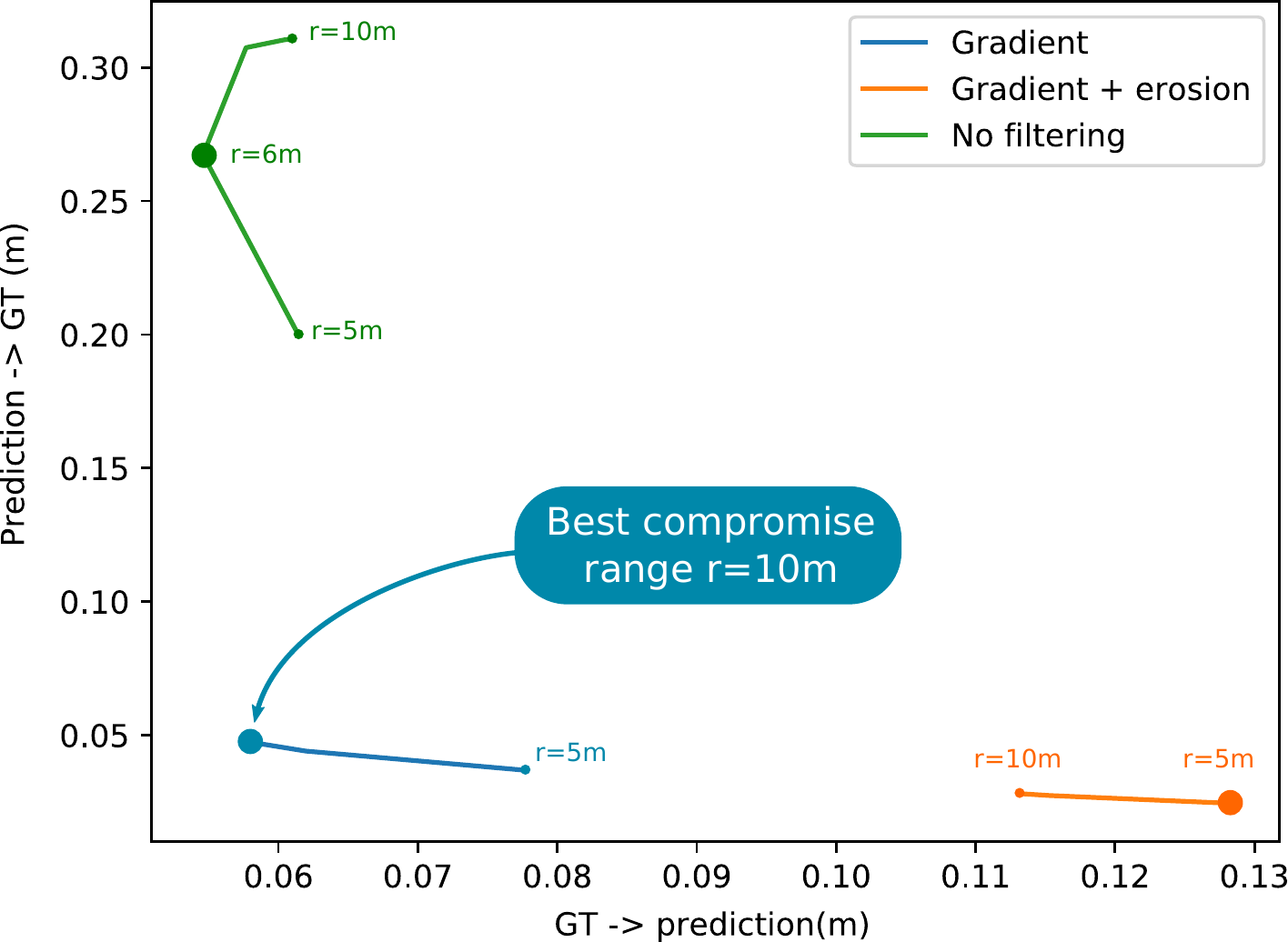}
\end{minipage}
\hfill
\begin{minipage}{0.48\linewidth}

\begin{minipage}{0.49\linewidth}
\centering
\includegraphics[width=\linewidth]{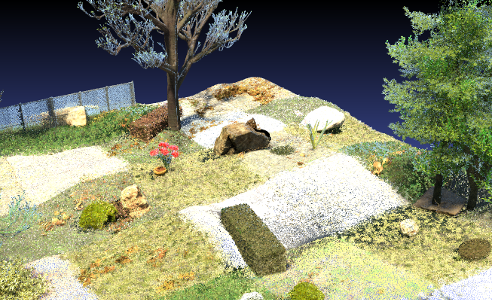}\\
Ground truth.
\end{minipage}
\hfill
\begin{minipage}{0.49\linewidth}
\centering
\includegraphics[width=\linewidth]{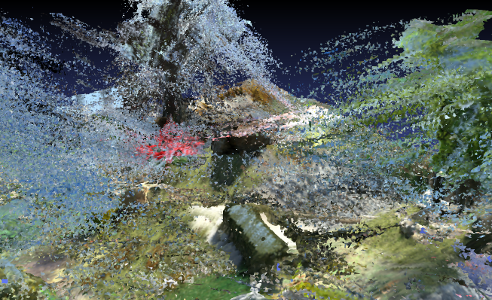}\\
No filter, r=6m.
\end{minipage}

\begin{minipage}{0.49\linewidth}
\centering
\includegraphics[width=\linewidth]{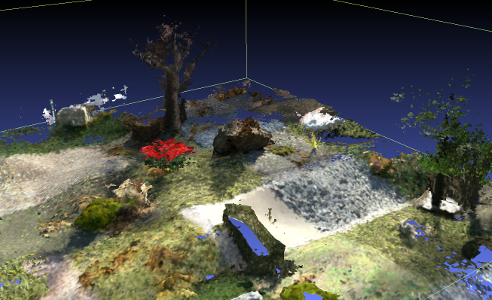}\\
Erosion, r=5m.
\end{minipage}
\hfill
\begin{minipage}{0.49\linewidth}
\centering
\includegraphics[width=\linewidth]{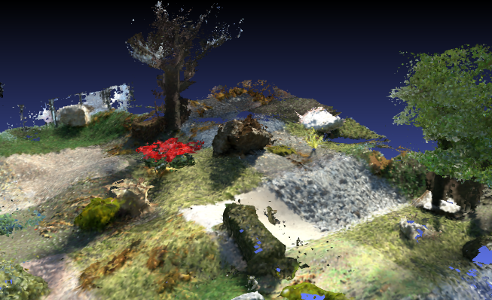}\\
Gradient, r=10m.
\end{minipage}

\end{minipage}
\caption{Influence of the clipping range and the filtering method on 3D reconstruction.}
\label{fig:3Deval}
\end{figure}

The compromise between completeness and accuracy is illustrated on figure~\ref{fig:3Deval}.
It presents, on the left, a graph showing the average distance for completeness function of the average distance for accuracy.
The curves represent the evolution of theses distance with respect to the clipping range.
The right part of the figure shows illustration for methods and ranges corresponding to the bigger dots in the previous graph.

\begin{figure}[t]
	\includegraphics[width=\linewidth]{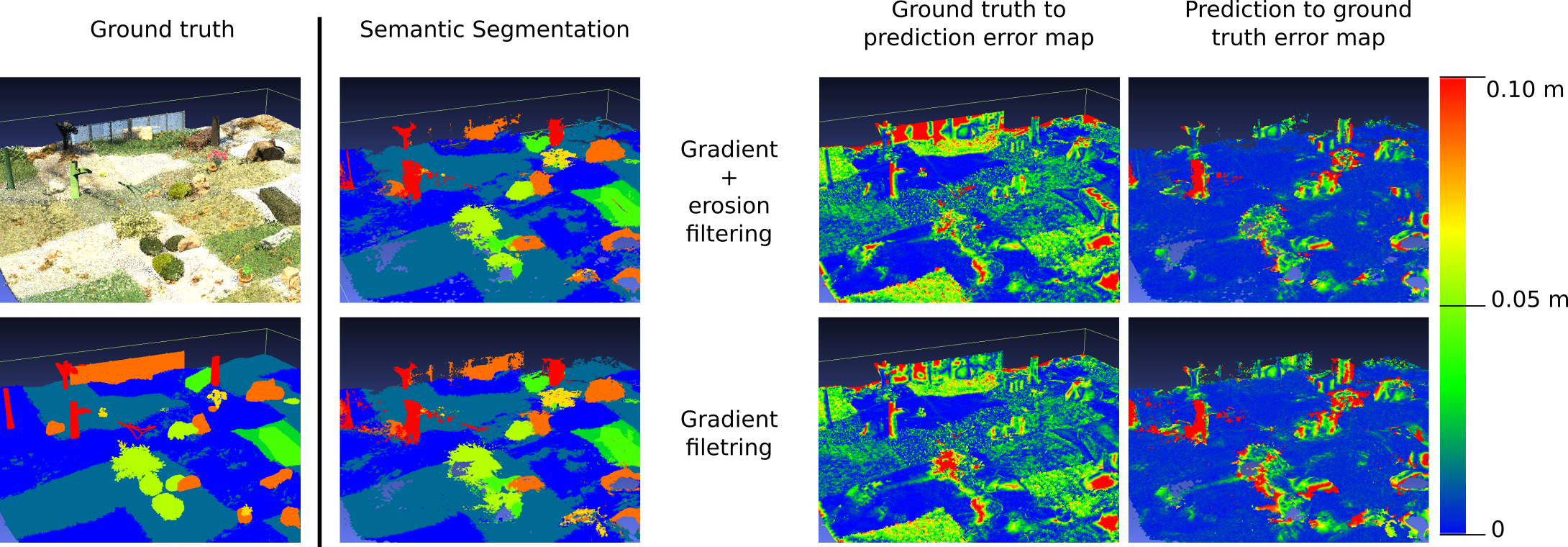}
    \caption{Reconstructions for two filtering policies: Semantics mesh (left) and error heat maps(right).}
    \label{fig:recons}
\end{figure}

Finally, error maps are presented on the left side of Fig.~\ref{fig:recons}.
For the \textit{GT} $\rightarrow$ \textit{Predictions} maps, the red points (error greater than 10cm) are the missing parts. For the \textit{Predictions} $\rightarrow$ \textit{GT} maps, red points correspond to  hallucinated objects, particularly multiple tree trunks or flowers.

\subsubsection{Semantics}~

Evaluation of 3D semantics is not straightforward: there is no direct correspondence between points of the ground truth and the reconstructed mesh.
We use a strategy to create a geometric clone of the ground truth and assigning to each point the label of the nearest vertex of our reconstructed mesh.
Thus, we have prediction/GT label pairs usable for metric computation.
Table~\ref{tab:3Dsem} presents the results for the \textit{Multi-task Net.} with gradient filtering for overall accuracy (OA), average accuracy (Av Acc.) and average intersection over union (Av. IoU).
Left side of figure~\ref{fig:recons} shows snapshots of the surface with semantics labels.
Most of the errors are located on the ground, mostly mixed grass and ground.
The semantic prediction tend to produce smooth segmentations and fail to create very small connected component, such as pebbles in the grass.

\begin{table}[t]
\caption{Evaluation of the 3D semantics.}
\label{tab:3Dsem}
\centering
\begin{tabular}{l@{\hspace{2mm}}l@{\hspace{2mm}}l@{\hspace{2mm}} c@{\hspace{2mm}}c@{\hspace{2mm}}c l@{\hspace{2mm}} c@{\hspace{2mm}}c@{\hspace{2mm}}c}
\toprule
                        &&& \multicolumn{3}{c}{Full scene} && \multicolumn{3}{c}{Cropped scene z=1m} \\
\cmidrule{4-6}\cmidrule{8-10}
Method  & range & Dataset & OA     & Av. Acc. & Av. IoU && OA     & Av. Acc. & Av. IoU\\
Gradient & 10m & 001 & 0.8950 & 0.8735 & 0.7285 && 0.8640 & 0.8705 & 0.7339 \\
\bottomrule
\end{tabular}
\end{table}

\begin{figure}[t]
\includegraphics[width=\linewidth]{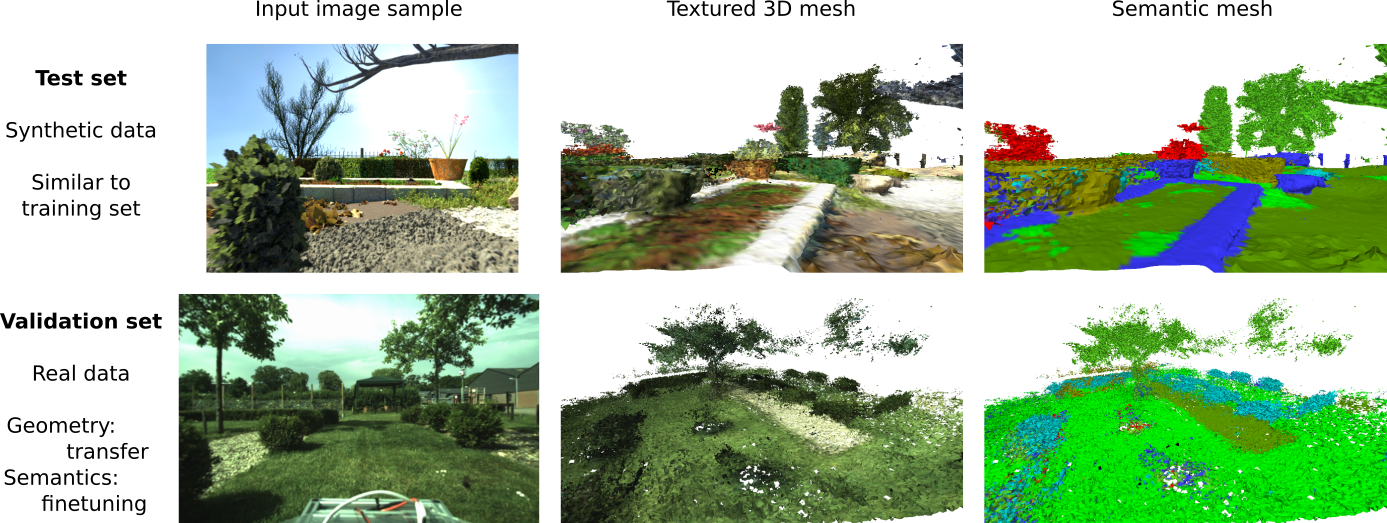}
\caption{Reconstructions on the tests sets of the \3drms datasets 2017 and 2018.}
\label{fig:transfer}
\end{figure}

\subsubsection{Transfer to real data}~

The ultimate goal of the reconstruction pipeline is to be applicable to real data.
To test the ability of our pipeline to generalize from the synthetic dataset to real outdoor data, we experiment the \3drms 2017 test dataset.
The results are shown on Fig.~\ref{fig:transfer}.
For visual comparison, we confront the reconstruction for the synthetic test (first row) and real data (second row), note the high difference between the two sample images.
We tested first direct transfer of the neural network to the new dataset.
While depth estimation was still efficient (middle image), semantic segmentation was deteriorated.
To address this problem, we finetuned the segmentation decoder of the network on the train set of the \3drms 2017 dataset for ten epochs.
Note that, in order to maintain the depth estimation quality, as the finetuning does not include depth ground truth, we froze the weights of the encoder and the depth decoder.
Results are the in the right column: the semantics of the main objects and ground classes are well recovered.

\subsection{Timings}~

We give some timing results for each separate block of our pipeline.  The experiments were carried with an Intel Xeon CPU E3-1505M and Nvidia GTX1070 GPU.
The stereo depth map estimation with SGBM takes 0.03s, the multi-task network depth and semantic inference takes 0.4s and the filtering step has a negligible computation time.  For the 3D reconstruction, with a clipping range of 5m and with resolutions of 3cm, 5cm and 10cm, the related run times are respectively 0.4s, 0.15s and 0.1s per depth map.  Higher clipping ranges increases the computation load as it significantly augment the number of points to use in the reconstruction.  With a range of 10m and a resolution of 3cm, the integration of one depth map takes 2.2s.
As our pipeline is designed to process incoming data online, one can expect each stereo pair to be processed in less than 0.85s for a high resolution map (3cm voxels - 5m range) and in less than 0.5s for lower resolution maps ($\geq 5$cm voxels - 5m range), which are often sufficient for autonomous navigation. 


\section{Conclusion}

In this paper, we have presented a 3D reconstruction approach from multiple stereo image pairs.
The reconstruction pipeline mixes both the accuracy of geometric approaches and the complex, high-order modeling made possible by  the deep neural networks.
We show co-learning of depth estimation and pixelwise semantic labeling is possible in robotics scenarios and improves the framework at every stage. Indeed,
the multi-task network, while being lighter than separate networks, is also more effective.
The proposed approach is compatible with online mapping and does not require global optimization.
The method has been evaluated on the \3drms 2018 simulated dataset.

A close look at the reconstructed surfaces shows that most of the geometric errors come from the duplication of some objects.
Moreover, the main part of semantic errors are due to mis-detected pixels which deteriorate the global score while most of the othe objects have been correctly recognized.
To improve these aspects of the method, future works will include performing object detection and tracking during the sequence. First, the object identification between images would reduce the number of instances in the final product and second, labels would be regularized at object level.
Even though it would make our method less suitable for online use, we also examine the possibility of a posteriori spatial regularization on the reconstructed surface such as conditional random fields.

\clearpage

\bibliographystyle{splncs}
\bibliography{biblio.bib}

\end{document}